# GLoT: A Novel Gated-Logarithmic Transformer for Efficient Sign Language Translation

Nada Shahin
*Intelligent Distributed Computing and Systems (INDUCE) Lab,
Department of Computer Science and Software Engineering,
College of Information Technology*
UAE University, Al-Ain, United Arab Emirates

Leila Ismail[1][2]
[1]*Intelligent Distributed Computing and Systems (INDUCE) Lab,
Department of Computer Science and Software Engineering,
College of Information Technology*
[2]*Emirates Center for Mobility Research*
UAE University, Al-Ain, United Arab Emirates
Correspondence: Leila@uaeu.ac.ae

*Abstract -* **Machine Translation has played a critical role in reducing language barriers, but its adaptation for Sign Language Machine Translation (SLMT) has been less explored. Existing works on SLMT mostly use the Transformer neural network which exhibits low performance due to the dynamic nature of the sign language. In this paper, we propose a novel Gated-Logarithmic Transformer (GLoT) that captures the long-term temporal dependencies of the sign language as a time-series data. We perform a comprehensive evaluation of GloT with the transformer and transformer-fusion models as a baseline, for Sign-to-Gloss-to-Text translation. Our results demonstrate that GLoT consistently outperforms the other models across all metrics. These findings underscore its potential to address the communication challenges faced by the Deaf and Hard of Hearing community.**

## I. INTRODUCTION

The number of Deaf and Hard of Hearing (DHH) population is expected to double to 860 million by 2050 [1]. In addition to the existence of more than hundreds of sign languages [2], and an acute shortage of sign language interpreters [3], there is a pressing need for automated and precise Sign Language Machine Translation (SLMT) systems. Having an inclusive communication could be lifesaving in a tragic event such as a medical emergency. Furthermore, in the era of smart cities health and pleasant experiences are essential [4], [5]. Consequently, developing a precise and efficient real-time SLMT system is crucial.

Machine Translation (MT) has been foundational in reducing language barriers, enabling seamless communication across diverse communities since the 1950s [6]. With the evolution of computational models, particularly the introduction of the transformer [7], MT has seen unparalleled advancements in accuracy for sequential data processing, such as spoken translation systems [8]. However, MT still facing challenges in language context complexity [9] and idiomatic expressions [10].

In recent years, the concept of applying MT techniques to convert sign language into spoken language and vice versa, known as SLMT systems, has gained increasing attention. SLMT systems consist of two parts: 1) Sign Language Recognition (SLR) and 2) Sign Language Translation (SLT). SLR, also known as sign-to-gloss (S2G) translation, transforms sign language videos into glosses, a written representation of sign language. This process is divided into 1) isolated sign language recognition (ISLR), which recognizes individual signs word by word without grammatical context, and 2) continuous sign language recognition (CSLR), which recognizes full sentences while considering grammatical structure. On the other hand, SLT translates sign language videos into text, either directly (S2T) or by incorporating gloss into the process (S2G2T). Using gloss as an intermediate step in S2G2T leads to more precise translations compared to the direct S2T method [11]. Therefore, in this paper, we focus on S2G2T using the transformer which is the most common and precise model in SLMT systems [12]. However, existing transformer models do not capture the long-term dependencies of the signs [13] which are of a temporal nature [14]. Each gesture, hand movement, facial expression, and transition between signs is sequential, creating a continuous and dynamic flow of information. SLMT systems must accurately capture both short-term dependencies, like the movement transitions of a sign, and long-term dependencies, such as the context in sentences [15]. In this paper, we aim to address this void by proposing the Gated Logarithmic Transformer (GLoT), which introduces a gating mechanism that selectively filters out irrelevant information, ensuring that only the most critical temporal dependencies are retained [16]. By incorporating logarithmic transformations, GLoT is designed to better capture long-range temporal patterns, improving the model's ability to learn sequential dependencies. Additionally, GLoT employs a cross-stage partial network, which enhances the flow of information across layers for better handling of long sequences in time-series data. These innovations allow GLoT to perform more efficiently in S2G2T translation, compared to the original transformer.

The main contributions of this paper are as follows:
- We propose a novel Gated-Logarithmic Transformer (GLoT) to enhance the S2G2T translation.
- We evaluate the performance of GLoT compared to the most precise transformer model in the literature and the transformer-fusion model baselines.
- We run the experiment using two continuous sign language datasets.
- We create a novel continuous medical sign language dataset as a testbed for our experiments.

The rest of the paper is organized as follows: Section II presents the related works. Section III describes the methodology. Section IV shows the experiments. Section V discusses the results and numerical analysis. Section VI concludes the paper. Lastly, Section VII presents the limitations.

## II. RELATED WORKS

Table I presents related works on S2G2T SLMT [11], [17], [18], [19], [20], [21], [22], [23]. Multiple algorithms were investigated, such as the transformer [11], [17], [18], [19], Convolution Neural Networks (CNN) [20], Long-Short Term Memory (LSTM) [21], Graph Neural Networks (GNN) [22], and Gated Recurrent Networks (GRU) [23]. These algorithms were validated on several datasets such as the Chinese datasets: CSL-Daily and CSL, and the German datasets: RWTH-PHOENIX-Weather-2014 (PHOENIX-2014T) and PHOENIX-2014. The table shows that the transformer is the most precise [11] and PHOENIX-2014T dataset [23] is the mostly used. Nevertheless, existing transformers in the literature do not consider the long-term temporal dependencies of the signs. Therefore, in this work, we propose GLoT which considers the time-series nature of the sign language by incorporating two novel mechanisms: the Stacked LogSparse Self-Attention (LSSA), which reduces computational complexity by focusing only on log-spaced past frames, and a Gating Mechanism that selectively enhances long-term dependencies while filtering irrelevant information. This allows GLoT to handle both short-term and long-term temporal patterns more effectively than existing models.

## III. METHODOLOGY

### A. Problem Definition

In this work, we consider the SLT problem that takes sign videos with $F$ frames $x = \{x_f\}_{f=1}^{F}$ as input and translates it into a series of spoken language text with $T$ sequence length as an output $y = \{y_t\}_{t=1}^{T}$. Our goal is to learn the mapping between sign language and spoken language for real-time translation using our proposed GLoT, depicted in Fig. 1.

### B. GLoT Architecture

We propose GLoT, a novel variation of the transformer encoder that includes the following components: a convolution layer, $LSSA$ [24], and Gating Mechanisms [16]. We describe the complete GLoT components below:

   a) Input Dimension: The sign videos $x$ are the encoder input and have a dimension of $m \times n$ that we flatten before feeding it to the encoder. The predicted gloss ($G'$)

TABLE I. COMPARISON BETWEEN SIGN LANGUAGE MACHINE TRANSLATION IN LITERATURE

| Work | Sign Language(s) | Dataset(s) | Algorithm(s) | BLEU-4 |
|---|---|---|---|---|
| [11] | Chinese and German | CSL-Daily and PHOENIX-2014* | Transformer | 23.51 |
| [17] | German | PHOENIX-2014T | Transformer | 23.1 |
| [18] | German | PHOENIX-2014T | Transformer | 22.45 |
| [19] | Chinese and German | CSL-Daily* and PHOENIX-2014T | Transformer with Selective Mutual knowledge Distillation | 23.8 |
| [20] | German and Chinese | CSL-Daily and PHOENIX-2014T* | Separable 3D CNN with mBart | 24.60 |
| [21] | Chinese and German | CSL, PHOENIX-2014 and PHOENIX-2014T* | BiLSTM | 23.7 |
| [22] | Chinese and German | CSL-Daily and PHOENIX-2014T* | Hierarchical GNN | 22.3 |
| [23] | German | PHOENIX-2014T | GRU with attention layer | 18.1 |

*: Best performance

and predicted text ($T'$), with the dimension $b$ and $d$ respectively, are the decoder inputs that we also embed before entering the positional encoding as well. However, $T'$ is an input to the decoder during training only.

   b) Encoder: We split the sign video input ($x_e$) into two halves $x_e = [x_{e1}^{F \times m_1 \times n_1}, x_{e2}^{F \times m_2 \times n_2}]$. $x_{e1}$ is passed through a convolution layer ($Conv$) to extract local features from the video frames, such as hand shapes and facial expressions. $x_{e2} = [x_{e2LSSA}, x_{e2GAP}]$. $x_{e2LSSA}$ is processed by a stacked $LSSA$ and $x_{e2GAP}$ is passed through Global Average Pooling ($GAP$), which is followed by a gating mechanism [16]. The final outputs of all input splits are then concatenated and added to $x_e$ and then normalized. The encoder output is given by Equation (1).

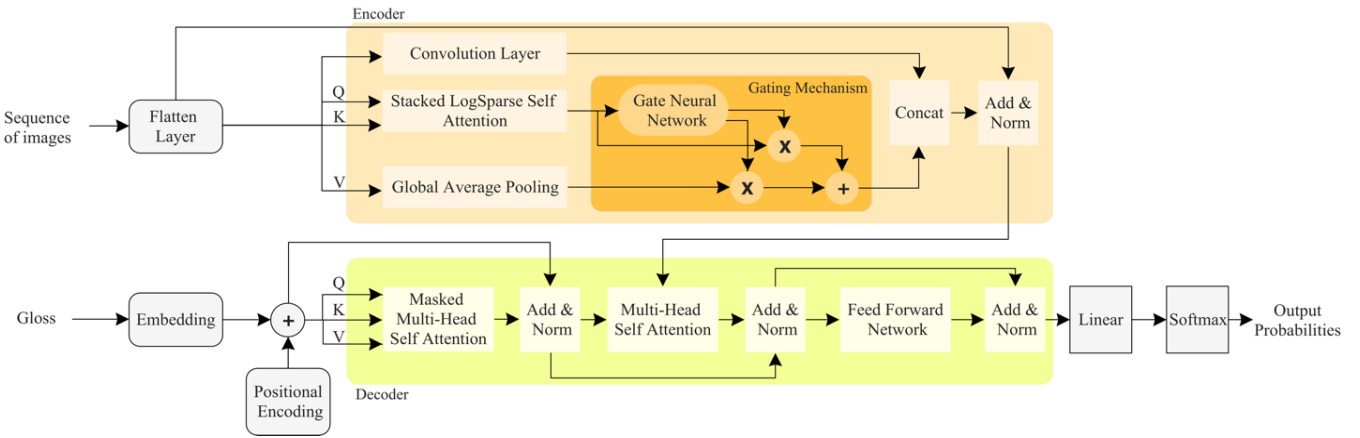

Fig 1. Proposed Transformer.

$$y_e = Concat(Conv(x_{e1}), (Gating(LSSA(x_{e2}), GAP(x_{e2})))) \quad (1)$$

- Stacked *LSSA*: Unlike the self-attention mechanism proposed by [7], where the attention scores are calculated between every pair of patches in $x$ with the complexity of $O(L^2)$, *LSSA* [24], computes the attention score for a logarithmic subset of previous patches in $x_{e2}$ with exponentially increasing step size. This reduces the number of calculations and the computational complexity to $O(L(logL)^2)$ while capturing long-term dependencies. In [7] and [24], the input is split into query, key, and value (Q/K/V) matrices. In contrast, in our proposed encoder, we only consider the query and key (Q/K) matrices. However, similar to [24] we employ stacked *LSSA* layers to ensure that all information in $x_{e2LSSA}$ is captured. The formula for LSSA is given in Equation (2).

$$LSSA(Q, K) = Sofmax\left(\frac{QK^T_{I^j_p}}{\sqrt{\frac{d}{2}}}\right) \quad (2)$$

where $d$ is the dimension and the set $I^j_p$ includes the indices of the patches that the current patch $p$ can attend to during the computation from $j$ to $J+1$. In other words, $I^j_p = \{p - 2^{\lfloor log_2 p \rfloor}, p - 2^{\lfloor log_2 p \rfloor - 1}, ..., p - 2^0, p\}$.

- *GAP*: The V matrix from the Q/K/V goes through 1D *GAP* to enhance position sensitivity within the encoder. This pooling operation computes the average value of all elements in V, summarizing the information in $x_{e2}$.
- Gating Mechanism [16] combines the outputs of the *LSSA* and *GAP* and takes them as inputs. This mechanism consists of a gate neural network that produces a gate value ($g$) which evaluates the attention weights and decides, whether to consider the output of *LSSA* or *GAP*. This is to decide the relative importance of short-term and long-term information, selectively enhancing important temporal dependencies, and to improve sensitivity to the relative positions of $x_{e2}$. The formula of this gating mechanism is given in Equations (3) and (4).

$$g = w \cdot LSSA + b \quad (3)$$

$$Gating = g \cdot LSSA + (1-g) \cdot GAP(V) \quad (4)$$

where $w$ is a random weight and $b$ is the bias.

- Positional Encoding (PE) learns the spatial relationships $T'$ and $G'$.
- Decoder Components: The decoder in our proposed GLoT has the same structure as the original transformer.

Together, these components capture both local details and long-range dependencies, improving the overall translation quality and complexity by focusing on the dynamic and temporal nature of sign language.

## IV. EXPERIMENTAL SETUP

### A. Datasets

- **PHOENIX-2014T** [23] is a multi-signer dataset collected from a German weather forecast broadcast over three years. It features RGB German Sign Language (DGS) videos with 8 signers. The dataset contains 8,247 videos with a resolution of 210×260 and a vocabulary size of 1,085 glosses. These videos are divided into 7,096 training instances, 519 validation instances, and 642 testing instances. In this work, we use a subset of this dataset that consists of 500 signed videos with a similar ratio of the training, validation, and testing instances.
- **MedASL** is our private medical-related single-signer dataset that consists of 500 signed videos with a resolution of 1280×800 and a vocabulary size of 832 glosses.

### B. Implementation Details

To develop an efficient SLMT system based on our proposed transformer architecture GLoT, we implement the hyperparameters presented in Table II. These hyperparameters yielded the best results on the transformer [11] and transformer-fusion [25], which serve as baselines for evaluating our proposed transformer. We use a 5-fold cross-validation method for training and validation, and report results based on the best-performing fold, which is subsequently used for testing. While the PHOENIX-2014T dataset is pre-divided into training, validation, and testing sets, we split the MedASL dataset into 80% for cross-validation and 20% for testing. We conduct all experiments using Python 3.10 on Supermicro with 1024 GB RAM, 2 AMD EPYC 7552 48-Core, 96-Thread, 2.20 GHz Processors, and 2 x NVIDIA RTX A6000 GPUs.

TABLE II. HYPERPARAMETERS FOR THE ALGORITHMS UNDERSTUDY

| Hyperparameter | Values used in literature | Values used in our experiments | |
|---|---|---|---|
| | | *Set 1 [11]* | *Set 2 [25]* |
| # encoders | NR [11], [25] | 1[*] | 1[*] |
| # decoders | NR [11], [25] | 1[*] | 1[*] |
| Hidden Units | 256 [25], 512 [11] | 512 | 256 |
| # Heads | 8 [11], [25] | 8 | 8 |
| Feedforward size | 2048 [11], 256 [25] | 2048 | 256 |
| Dropout | 0.1 [11], No dropout [25] | 0.1 | No dropout |
| Learning rate | $5 \times 10^{-5}$ reduced by a factor of 0.5 until $2 \times 10^{-6}$ for 3 steps [11], 0.001 [25] | $5 \times 10^{-5}$ reduced by a factor of 0.5 until $2 \times 10^{-6}$ for 3 steps | 0.001 |
| Epochs | 32 [25], NR [11] | 30 | 30 [11] |
| Batch size | 32 [25], NR [11] | 32 | 32 [11] |

NR: Not reported. [*]: Default value

## C. Evaluation Metric

We use the Bilingual Evaluation Understudy (BLEU) metric [26] to measure the similarity between the machine and human translations. The BLEU scores are presented between 0 and 1, where 1 is the most precise match. These scores also consider the n-gram from 1 to 4 (BLEU-1 to BLEU-4). Equations 5 and 6 show the calculation of the BLEU score.

$$BLEU = BP \cdot e^{(\sum_{n=1}^{N} w_n \log(p_n))} \quad (5)$$

where $p_n$ is the precision of n-grams, calculated as the ratio of the number of matching n-grams in the translation to the total number of n-grams in the translation, $w_n$ is the weight of each n-gram size, and $BP$ is the Brevity Penalty.

$$BP = \begin{cases} 1, & if\ c > r \\ e^{(1-\frac{r}{c})}, & if\ c \leq r \end{cases} \quad (6)$$

where $c$ is the length of the candidate machine translation and $r$ is the reference corpus length.

## V. RESULTS AND NUMERICAL ANALYSIS

Table III presents the performance evaluation of three transformer architectures—the original transformer, transformer-fusion, and our proposed GLoT —using PHOENIX-2014T subset and MedASL datasets. The results show that GLoT consistently outperforms the other transformer architectures across both datasets. In particular, GLoT achieves a BLEU-4 score of 0.067 versus 0.064 for the original transformer and 0 for transformer-fusion, using a subset of PHOENIX-2014T. It also scores a BLEU-4 of 0.085, versus 0.059 for the original transformer and 0.001 for transformer-fusion, using MedASL dataset. The improvement of GLoT is due to the long-term dependency between the signs. On the other hand, the poor performance of transformer-fusion across both datasets is due to the omission of normalization layers.

In summary, our proposed GLoT consistently delivers superior performance compared to the transformer and transformer-fusion across both datasets and various settings. Its ability to maintain higher BLEU scores underscores its effectiveness at capturing the linguistic nuances of sign language translation, while the near-zero performance of the fusion transformer highlights the challenges of using such architectures in this domain.

## VI. CONCLUSION

This paper presents a comprehensive evaluation of three transformer architectures— the transformer, transformer-fusion, and our proposed Gated-Logarithmic Transformer (GLoT)—for the task of Sign-to-Gloss-to-Text (S2G2T) translation. Our findings underscore the effectiveness of GLoT, which consistently outperforms both the transformer and transformer-fusion architectures across the PHOENIX-2014T and MedASL datasets. The proposed architecture not only delivers higher BLEU scores but also demonstrates a more reliable generalization from validation to testing phases.

The improvements we observed are particularly critical when considering the real-world implications of Sign Language Machine Translation (SLMT) systems. Sign language, unlike spoken language, involves intricate spatio-temporal patterns that incorporate hand movements, facial expressions, and body posture, which are crucial for semantic accuracy. GLoT's ability to capture these multimodal nuances more effectively than other models suggests its potential for broader applications, including real-time translation in various domains, such as education, healthcare, and emergency services.

Moreover, the strong performance of our model on the MedASL dataset, which consists of medical terminology that requires precision, demonstrates the viability of using this model in specialized fields where communication accuracy is paramount. This is particularly important in environments where miscommunication could lead to dire consequences, such as in medical or legal contexts. The flexibility and adaptability of GLoT emphasize its broader applicability across different sign languages and contexts, potentially mitigating the shortage of qualified sign language interpreters.

TABLE III.  COMPARISON BETWEEN TRANSFORMER ARCHITECTURES WITH TWO HYPERPARAMETER CONFIGURATIONS (SETS) ON THE PHOENIX-2014T SUBSET AND MedASL DATASETS

| Dataset | Architecture (hyperparameter set) | Validation | | | | Testing | | | |
|---|---|---|---|---|---|---|---|---|---|
| | | *BLEU-1* | *BLEU-2* | *BLEU-3* | *BLEU-4* | *BLEU-1* | *BLEU-2* | *BLEU-3* | *BLEU-4* |
| PHOENIX-2014T subset | Original transformer (set 1) | 0.347 | 0.195 | 0.121 | 0.077 | 0.247 | 0.126 | 0.085 | 0.064 |
| | Original transformer (set 2) | 0.634 | 0.562 | 0.525 | 0.495 | 0.272 | 0.135 | 0.085 | 0.057 |
| | Transformer-fusion (set 1) | 0 | 0 | 0 | 0 | 0 | 0 | 0 | 0 |
| | Transformer-fusion (set 2) | 1.308E-12 | 8.306E-13 | 6.063E-13 | 4.032E-13 | 0 | 0 | 0 | 0 |
| | **GLoT (set 1)** | **0.369** | **0.217** | **0.144** | **0.096** | **0.273** | **0.139** | **0.092** | **0.067** |
| | GLoT (set 2) | 0.494 | 0.324 | 0.232 | 0.166 | 0.281 | 0.137 | 0.081 | 0.057 |
| MedASL | Original transformer (set 1) | 0.442 | 0.308 | 0.226 | 0.163 | 0.287 | 0.177 | 0.107 | 0.062 |
| | Original transformer (set 2) | 0.538 | 0.429 | 0.361 | 0.299 | 0.272 | 0.159 | 0.098 | 0.059 |
| | Transformer-fusion (set 1) | 0 | 0 | 0 | 0 | 0 | 0 | 0 | 0 |
| | Transformer-fusion (set 2) | 0.041 | 0.006 | 0.003 | 0.002 | 0.034 | 0.006 | 0.002 | 0.001 |
| | GLoT (set 1) | 0.439 | 0.294 | 0.211 | 0.148 | 0.287 | 0.178 | 0.105 | 0.064 |
| | **GLoT (set 2)** | **0.580** | **0.486** | **0.421** | **0.358** | **0.323** | **0.203** | **0.133** | **0.085** |

In summary, our proposed model represents a significant step forward in SLMT, offering an efficient and accurate solution for S2G2T translation. Its success provides valuable insights into the architectural elements necessary for future advancements, underscoring the importance of domain-specific adaptation and scalability in addressing the communication challenges faced by the global Deaf and hard-of-hearing community.

## VII. Limitations and future work

While our proposed transformer, GLoT, shows significant improvements over existing architectures, we acknowledge several limitations and areas for future work. First, the evaluation was conducted using datasets that may not fully represent the diversity of real-world sign language usage across different sign languages and contexts. The PHOENIX-2014T dataset primarily consists of weather-related content, limiting the model's exposure to more varied and complex sentence structures. Similarly, the MedASL dataset, while focused on medical terminology, is a single-signer dataset that does not account for the variation in signing styles seen in broader multi-signer scenarios. To strengthen the findings and demonstrate GLoT's generalizability, future work should expand the evaluation to include larger, more diverse datasets such as those covering everyday conversation, formal settings, and multiple sign languages. Second, while GLoT outperformed the other architectures, the improvement in BLEU-4 scores, particularly on the PHOENIX-2014T subset, was minimal. This suggests that there is still room for optimization, particularly in handling more complex sign language structures, such as capturing sign language non-manual markers which include facial expressions and body language [12]. These can be addressed by incorporating multimodal data and feature extraction techniques [27]. Third, pre-training GLoT on larger, more diverse datasets could expose the model to broader linguistics. This will allow it to learn richer sign representations and generalize better across sign languages and domains. Hence, the translation quality in real-world applications would improve. In addition, we will analyze the complexity of GLoT versus the original transformer. Training time in transformer architectures remains a concern, as this limits their real-time accessibility in real-life scenarios. Finally, future work should explore optimization techniques that reduce the computational overhead while maintaining the translation performance. It includes model compression techniques such as model quantization [28] and knowledge distillation [29] to reduce the deep learning models' size and computational costs. In addition, deploying a lightweight GLoT on edge devices could make real-time sign language translation more accessible [30].

## Acknowledgment

This work was supported by the United Arab Emirates University.